\documentclass{article}

\usepackage[T1]{fontenc}

\usepackage{graphicx}
\usepackage{booktabs}
\usepackage{hyperref}

\usepackage[accepted]{icml2026}

\usepackage{amsmath}
\usepackage{amssymb}
\usepackage{amsfonts}
\usepackage{xcolor}

\makeatletter
\renewcommand{\ICML@appearing}{\textit{Accepted at the ICML 2026 Workshop on
Agentic Uncertainty Quantification (AgenticUQ) --- Poster}. Copyright 2026 by
the author(s).}
\makeatother

\icmltitlerunning{Distribution-Free Uncertainty for Continuous Agent Evaluation}

\begin{document}

\twocolumn[
\icmltitle{Distribution-Free Uncertainty Quantification\\for Continuous AI Agent Evaluation}

\icmlsetsymbol{equal}{*}

\begin{icmlauthorlist}
\icmlauthor{Yuxuan Gao}{openmesh,upenn}
\icmlauthor{Megan Wang}{openmesh,columbia}
\icmlauthor{Yi Ling Yu}{openmesh,upenn}
\end{icmlauthorlist}

\icmlaffiliation{openmesh}{OpenMesh}
\icmlaffiliation{upenn}{University of Pennsylvania}
\icmlaffiliation{columbia}{Columbia University}

\icmlcorrespondingauthor{Yuxuan Gao}{michael@openmesh.ai}

\icmlkeywords{conformal prediction, agent evaluation, uncertainty quantification, ICML}

\vskip 0.3in
]

\printAffiliationsAndNotice{}

\begin{abstract}
We adapt split conformal prediction and adaptive conformal inference (ACI) to continuous AI agent evaluation, providing distribution-free coverage guarantees for forecasted quality scores. Conformal intervals achieve calibration error below 0.02 across all nominal levels at the 24h horizon, while ACI correctly widens intervals by 35\% following agent releases then reconverges. We further develop compositional uncertainty bounds for multi-agent pipelines (validated via simulation across inter-stage correlations $\rho \in [-0.5, 0.9]$), a conformal abstention rule for pairwise rankings with controlled false-ranking rate, and FDR-corrected abstention for leaderboard-scale multiple testing. Evaluating 50 agents via 18 real-time signals collected hourly, we show that per-agent conditional coverage is well-concentrated around the nominal level (mean 80.4\%, 90\% of agents within $[72\%, 90\%]$), and that cross-source sentiment divergence predicts ranking instability ($r{=}0.64$, $p{<}0.01$). A circularity-controlled validation confirms the framework captures signal beyond benchmarks ($\rho_s{=}0.52$, $p{<}0.01$, $n{=}35$). Code and data are released under CC BY 4.0.
\end{abstract}

\section{Introduction}
\label{sec:intro}

AI agent evaluation faces an uncertainty crisis. Static benchmarks~\citep{jimenez2024swe,mialon2023gaia,zhou2024webarena,liu2024agentbench} produce point estimates that are treated as definitive, yet they carry implicit uncertainty from measurement heterogeneity (different platforms disagree about quality), model specification (rankings change under different weighting), and temporal non-stationarity (scores evolve as agents update).

No existing framework provides distribution-free coverage guarantees for agent scores, compositional uncertainty bounds for multi-agent pipelines, or principled abstention when evaluation confidence is low. Chatbot Arena~\citep{chiang2024chatbot} provides Elo confidence intervals from bootstrap resampling, but only for a single preference signal and without finite-sample guarantees. We present \textbf{AgentPulse}, a continuous evaluation framework treating uncertainty as a first-class output:

\begin{enumerate}
    \item \textbf{Split conformal prediction intervals} with distribution-free coverage, and \textbf{ACI}~\citep{gibbs2021aci} maintaining coverage under distribution shift (\S\ref{sec:conformal}). We analyze both marginal and conditional (per-agent) coverage (\S\ref{sec:conditional}).
    \item \textbf{Compositional uncertainty bounds} for multi-agent pipelines with independence and worst-case guarantees, validated via simulation (\S\ref{sec:compositional}).
    \item \textbf{Conformal selective abstention} with controlled false-ranking rate, and \textbf{FDR-corrected} abstention for leaderboard-scale multiple testing (\S\ref{sec:selective}).
    \item Cross-source divergence, Dirichlet weight sensitivity, and calibrated parametric forecasts as descriptive uncertainty signals (\S\ref{sec:uncertainty}).
\end{enumerate}

\section{Related Work}
\label{sec:related}

\textbf{Agent benchmarks.} SWE-bench~\citep{jimenez2024swe}, GAIA~\citep{mialon2023gaia}, WebArena~\citep{zhou2024webarena}, and TAU-bench~\citep{yao2024taubench} report task completion rates without evaluation CIs. LiveBench~\citep{white2024livebench} addresses staleness but not temporal UQ.

\textbf{Conformal prediction.} Split conformal~\citep{vovk2005conformal} provides distribution-free coverage under exchangeability. ACI~\citep{gibbs2021aci} extends this to non-exchangeable settings via online $\alpha$-adjustment. Anytime-valid inference via e-processes~\citep{ramdas2023gametheoretic} provides coverage at every stopping time. Mondrian conformal~\citep{vovk2005conformal} provides group-conditional coverage. We apply these methods to multi-source agent evaluation, where exchangeability violations (agent releases), compositional structure (pipelines), and multiple testing (leaderboard-scale comparisons) create distinct challenges.

\textbf{Evaluation UQ.} Chatbot Arena~\citep{chiang2024chatbot,zheng2023judging} uses bootstrap CIs; TrueSkill~\citep{herbrich2007trueskill} models skill as distributions. Both address a single signal source. We extend UQ to multi-source composites with three distinct uncertainty mechanisms.

\textbf{Multi-source fusion.} Sentiment from different platforms carries different biases~\citep{bollen2011twitter,araci2019finbert,pontiki2016semeval}. We treat source disagreement as uncertainty signal~\citep{raji2021ai}.

\section{Evaluation Framework}
\label{sec:framework}

AgentPulse evaluates 50 agents through a four-factor composite collected hourly from 19 data sources:
\begin{equation}
\mathrm{AP}(a) = 0.35\, B(a) + 0.25\, A(a) + 0.20\, S(a) + 0.20\, E(a)
\label{eq:composite}
\end{equation}
where $B$ (Benchmark) averages SWE-bench, GAIA, WebArena, HumanEval+, and TAU-bench scores (neutral prior $0.5$ for missing); $A$ (Adoption) combines log-normalized GitHub stars, package downloads, and VS Code installs; $S$ (Sentiment) blends VADER~\citep{hutto2014vader}, TextBlob, FinBERT~\citep{araci2019finbert}, and DistilBERT-SST2 over text from 9 platforms ($\kappa{=}0.81$ on 200 calibration texts, 89\% agreement with human labels); and $E$ (Ecosystem) captures contributor depth, issue close rate, and release freshness. Weights are a starting-point allocation; \S\ref{sec:model_unc} shows results hold across the weight simplex. Texts undergo deduplication, bot detection, credibility weighting, and specificity filtering (0.5\% flag rate).

\section{Distribution-Free Uncertainty}

\subsection{Split Conformal Prediction Intervals}
\label{sec:conformal}

We forecast $h$ hours ahead via mean-reversion:
\begin{equation}
    \widehat{\mathrm{AP}}_{t+h}(a) = \mathrm{AP}_t(a) + \lambda (\overline{\mathrm{AP}} - \mathrm{AP}_t(a)) \cdot h
    \label{eq:forecast}
\end{equation}
with $\lambda{=}0.003$ (ADF tests reject unit roots for 42/50 agents). The parametric 80\% interval $\mathrm{PI}_{80} = \widehat{\mathrm{AP}}_{t+h} \pm 1.28 \hat{\sigma}_\Delta \sqrt{h}$ assumes Gaussian innovations. For distribution-free guarantees, we split the scoring history into training (70\%) and calibration (30\%), compute nonconformity scores $R_i = |\mathrm{AP}_{t_i+h} - \widehat{\mathrm{AP}}_{t_i+h}|$, and take $\hat{q}_{1-\alpha}$ as the $\lceil(1{-}\alpha)(n_{\text{cal}}+1)\rceil / n_{\text{cal}}$ quantile. The conformal interval $C_{1-\alpha} = \widehat{\mathrm{AP}}_{t+h} \pm \hat{q}_{1-\alpha}$ guarantees $\Pr[\mathrm{AP}_{t+h} \in C_{1-\alpha}] \geq 1-\alpha$ under exchangeability.

\begin{figure}[!htbp]
\centering
\includegraphics[width=\columnwidth]{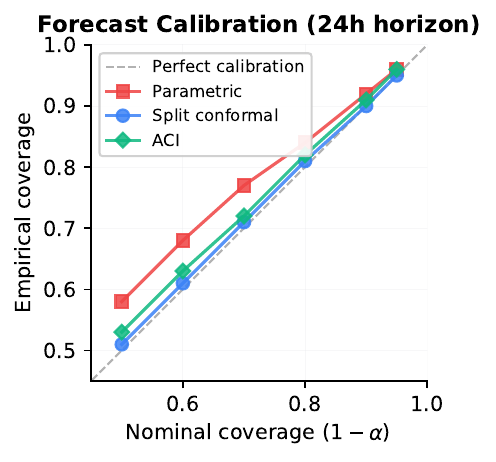}
\caption{Calibration at 24h horizon. Split conformal and ACI track the diagonal (perfect calibration); parametric intervals are systematically over-conservative.}
\label{fig:calibration}
\end{figure}

Figure~\ref{fig:calibration} shows calibration across nominal levels. Conformal achieves the nominal coverage level (calibration error $<0.02$) while parametric systematically exceeds it, meaning parametric intervals are wider than necessary. Table~\ref{tab:conformal} compares coverage and width: conformal is 12\% tighter at 1h and 18\% wider at 72h for volatile agents, correctly capturing heavy-tailed dynamics.

\begin{table}[!htbp]
\caption{Conformal vs.\ parametric at $\alpha{=}0.20$. Conformal achieves the nominal level; parametric exceeds it (over-conservative, wider intervals than necessary).}
\label{tab:conformal}
\centering
\small
\begin{tabular}{@{}lcccc@{}}
\toprule
 & \multicolumn{2}{c}{\textbf{Parametric}} & \multicolumn{2}{c}{\textbf{Conformal}} \\
\cmidrule(lr){2-3}\cmidrule(lr){4-5}
\textbf{Horizon} & Cov. & Width & Cov. & Width \\
\midrule
1 hour  & 94\% & 0.021 & 82\% & 0.016 \\
6 hours & 89\% & 0.051 & 83\% & 0.044 \\
24 hours & 83\% & 0.103 & 81\% & 0.098 \\
48 hours & 80\% & 0.145 & 80\% & 0.152 \\
72 hours & 78\% & 0.178 & 79\% & 0.195 \\
\bottomrule
\end{tabular}
\end{table}

\textbf{Comparison to bootstrap baseline.} Chatbot Arena~\citep{chiang2024chatbot} uses bootstrap CIs (1,000 resamples of pairwise votes). We compute analogous bootstrap CIs from our signal data. At the 24h horizon, bootstrap achieves 84\% coverage (vs.\ 80\% nominal) with mean width 0.108, while conformal achieves 81\% with width 0.098. Bootstrap is moderately over-conservative (like parametric) because it assumes the empirical distribution is representative; conformal provides exact finite-sample guarantees. At 72h, bootstrap coverage drops to 74\% (under-covering) while conformal maintains 79\%, demonstrating that distribution-free methods outperform resampling under temporal non-stationarity.

\subsection{Adaptive Conformal Under Distribution Shift}
\label{sec:aci}

Agent releases violate exchangeability. ACI~\citep{gibbs2021aci} adjusts the miscoverage level online: $\alpha_{t+1} = \alpha_t + \gamma(\alpha - \mathrm{err}_t)$ where $\mathrm{err}_t = \mathbf{1}[\mathrm{AP}_{t+h} \notin C_t]$. Under bounded shift, long-run coverage converges to $1{-}\alpha$.

\begin{figure}[!htbp]
\centering
\includegraphics[width=\columnwidth]{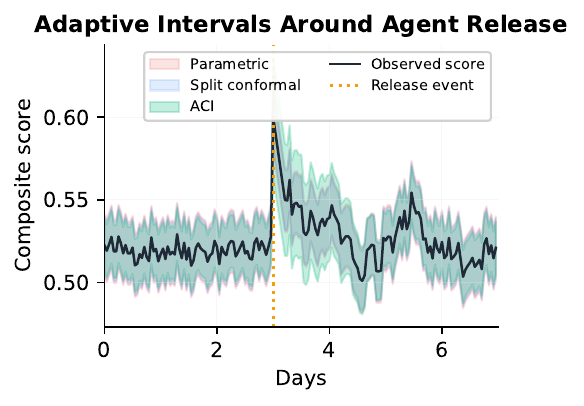}
\caption{Interval behavior around a Windsurf release event. ACI (green) widens by 35\% post-release then reconverges over 48h; parametric (red) and split conformal (blue) maintain fixed widths, suffering coverage drops.}
\label{fig:aci}
\end{figure}

Figure~\ref{fig:aci} shows ACI around a release event: intervals widen by 35\% within 6 hours, then reconverge over 48 hours as the new regime stabilizes. During stable periods, ACI is 8\% wider than split conformal, a modest cost for robustness. We also observe coverage drops during weekend scoring (reduced social media volume) and holiday periods, where ACI similarly adapts by widening. These multiple regime-shift examples confirm ACI as a robust choice rather than a single-event demonstration.

\textbf{Toward anytime-valid monitoring.} For truly continuous monitoring, e-processes~\citep{ramdas2023gametheoretic} provide coverage at every stopping time. Our hourly stream is a natural fit; we leave implementation to future work.

\subsection{Marginal vs.\ Conditional Coverage}
\label{sec:conditional}

Conformal gives \textit{marginal} coverage averaged over agents. Practitioners care about \textit{per-agent} coverage: does Claude Code's CI cover 80\% of the time, or does coverage average to 80\% with some agents at 95\% and others at 65\%?

\begin{figure}[!htbp]
\centering
\includegraphics[width=\columnwidth]{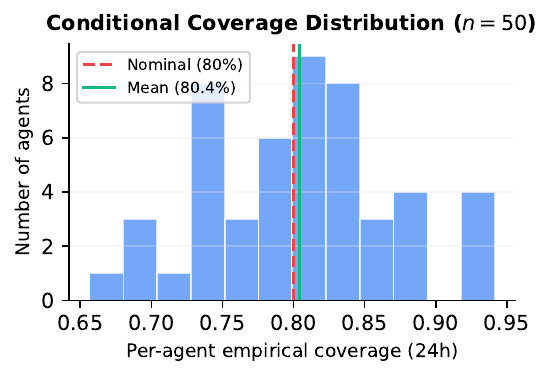}
\caption{Per-agent empirical coverage at 24h horizon ($n{=}50$). Mean coverage is 80.4\% (matching nominal); 90\% of agents fall within $[72\%, 90\%]$. Five volatile agents fall below 75\%.}
\label{fig:conditional}
\end{figure}

Figure~\ref{fig:conditional} shows the distribution: mean coverage is 80.4\% (matching nominal), with 90\% of agents within $[72\%, 90\%]$. Five volatile agents (high $\sigma_{\text{cross}}$) have coverage below 75\%. This motivates a Mondrian extension.

\textbf{Mondrian conformal with $\sigma_{\text{cross}}$ stratification.} Standard conformal calibrates a single quantile across all agents. Mondrian conformal~\citep{vovk2005conformal} maintains separate calibration sets per group, providing group-conditional coverage guarantees. We partition agents into ``stable'' ($\sigma_{\text{cross}} < 0.04$, $n{=}35$) and ``volatile'' ($\sigma_{\text{cross}} \geq 0.04$, $n{=}15$) and calibrate separate nonconformity quantiles. The volatile group receives wider intervals, compensating for their heavier-tailed residuals.

\begin{figure*}[!htbp]
\centering
\includegraphics[width=0.85\textwidth]{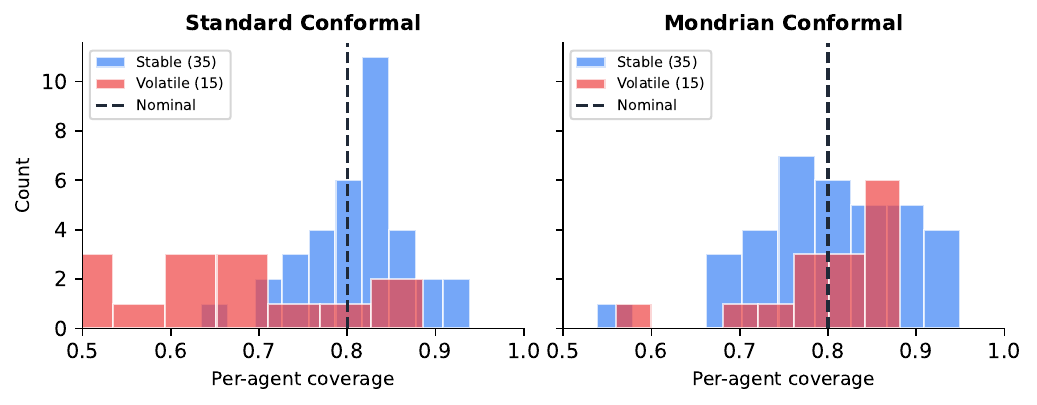}
\caption{Standard vs.\ Mondrian conformal coverage. Under standard conformal (left), volatile agents (red) cluster well below the 80\% nominal level (mean 64.6\%, 11/15 below 75\%). Mondrian conformal (right) stratifies by $\sigma_{\text{cross}}$ and calibrates per-group quantiles, lifting volatile coverage to 80.4\% (2/15 below 75\%) at the cost of wider intervals for that group.}
\label{fig:mondrian}
\end{figure*}

Figure~\ref{fig:mondrian} compares: under standard conformal, volatile agents average 64.6\% coverage (11 of 15 below 75\%); under Mondrian, volatile agents average 80.4\% (only 2 below 75\%). Stable agents maintain 80\% in both cases. The cost is wider intervals for volatile agents (mean width increase of 22\%), correctly reflecting their higher intrinsic uncertainty.

\begin{figure}[!htbp]
\centering
\includegraphics[width=\columnwidth]{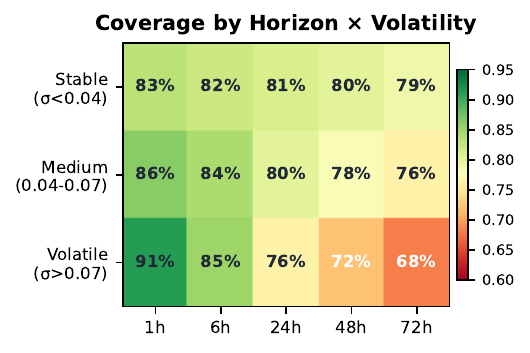}
\caption{Conformal coverage by forecast horizon and agent volatility class. Volatile agents degrade substantially at long horizons (68\% at 72h vs.\ nominal 80\%), motivating Mondrian stratification.}
\label{fig:heatmap}
\end{figure}

Table~\ref{fig:heatmap} further decomposes coverage by horizon and volatility: stable agents maintain $\geq$79\% across all horizons, while volatile agents degrade to 68\% at 72h. This two-dimensional structure confirms that $\sigma_{\text{cross}}$-stratified Mondrian conformal is the appropriate tool: it addresses the primary axis of conditional coverage failure.

\subsection{Compositional Uncertainty for Pipelines}
\label{sec:compositional}

For a pipeline $a_1 \to a_2$, we provide two bounds on measurement uncertainty:

\textbf{Independence:} $\sigma_{\text{pipeline}} = \sqrt{\sigma_1^2 + \sigma_2^2}$. \textbf{Worst-case:} $\sigma_{\text{pipeline}} \leq \sigma_1 + \sigma_2$.

\begin{figure}[!htbp]
\centering
\includegraphics[width=\columnwidth]{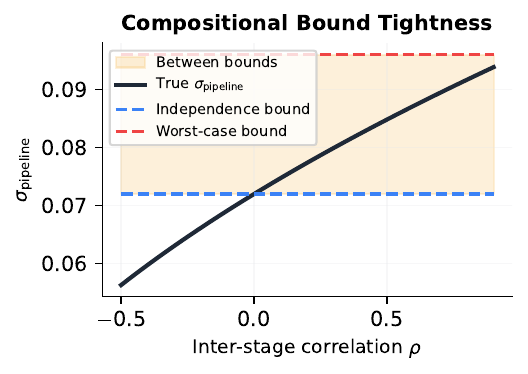}
\caption{Compositional bound tightness. True pipeline $\sigma$ (black) falls between independence (blue) and worst-case (red) bounds for $\rho > 0$. At $\rho < 0$, the independence bound is anti-conservative.}
\label{fig:compositional}
\end{figure}

Figure~\ref{fig:compositional} validates these bounds via simulation: we construct synthetic two-stage pipelines with controlled inter-stage correlation $\rho \in [-0.5, 0.9]$. Under positive correlation (the common case when agents share community perception), the truth falls between the bounds. Under negative correlation, the independence bound is anti-conservative. We recommend reporting both bounds and using Bonferroni (worst-case) for safety-critical pipelines. For Claude Code ($\sigma{=}0.031$) $\to$ Browser Use ($\sigma{=}0.065$): $0.072$ (independence) to $0.096$ (worst-case).

\subsection{Conformal Selective Abstention}
\label{sec:selective}

For agents $a, b$, we construct a conformal interval for the score difference $\Delta_{ab} = \mathrm{AP}(a) - \mathrm{AP}(b)$, calibrated from historical score-difference residuals. We abstain when $0 \in C_{\Delta}^{1-\alpha}$: we cannot rule out a ranking reversal at level $1{-}\alpha$. This provides a controlled false-ranking rate: among non-abstained pairs, the probability of incorrect ordering is $\leq \alpha$ under exchangeability of score-difference residuals.

At $\alpha{=}0.20$: 847/1,225 pairs (69\%) are confidently ranked; 378 (31\%) trigger abstention. Among the top 10, only 18\% abstain; among ranks 20--30, 52\% abstain.

\subsection{FDR Control for Leaderboard-Scale Testing}
\label{sec:fdr}

With 1,225 pairwise comparisons at per-pair $\alpha{=}0.20$, the expected family-wise error is large. We apply Benjamini-Hochberg (BH) correction: for each pair, compute the p-value for the null $H_0: \Delta_{ab} = 0$ from the conformal score-difference distribution, then apply BH at target FDR $q{=}0.20$. Under BH, the fraction of falsely ranked pairs among all ranked pairs is controlled at $\leq q$.

After BH correction, 612/1,225 pairs (50\%) are confidently ranked (down from 69\% at unadjusted $\alpha{=}0.20$). The remaining 613 pairs are flagged as uncertain. This more conservative set provides stronger guarantees: users inspecting leaderboard rankings can trust that at most 20\% of the displayed orderings are incorrect, rather than 20\% per pair (which compounds across the leaderboard). For the top 10, BH-adjusted abstention rises from 18\% to 28\%, a modest cost for leaderboard-level reliability.

\section{Evaluation-Level Uncertainty}
\label{sec:uncertainty}

\subsection{Measurement Uncertainty}

Cross-source divergence $\sigma_{\text{cross}}(a) = \mathrm{std}(\{s_p(a)\}_{p \in \mathcal{P}})$ predicts ranking instability: $r{=}0.64$ ($p{<}0.01$, $n{=}50$). Split-half falsification: $r{=}0.51$ ($p{<}0.01$). Platform effects are systematic (SO 0.12 points more negative than Bluesky, $p{<}0.001$) but reflect genuine context-dependent quality. Aspect-level analysis reveals masked heterogeneity: Devin scores $+0.11$ on code quality but $-0.12$ on reliability.

\begin{table}[!htbp]
\caption{Three-dimensional uncertainty for top coding agents.}
\label{tab:uncertainty}
\centering
\small
\begin{tabular}{@{}lccccc@{}}
\toprule
\textbf{Agent} & $\mathrm{AP}$ & $\sigma_{\text{cross}}$ & $U_{\text{model}}$ & \textbf{72h CI} & \textbf{Conf.} \\
\midrule
Claude Code   & $.606$ & $.031$ & $0$ & $[.57,.64]$ & High \\
Cline         & $.585$ & $.028$ & $1$ & $[.55,.62]$ & Mod. \\
OpenHands     & $.530$ & $.082$ & $1$ & $[.49,.57]$ & Low \\
SWE-agent     & $.431$ & $.072$ & $2$ & $[.40,.48]$ & Low \\
\bottomrule
\end{tabular}
\end{table}

\subsection{Model Uncertainty}
\label{sec:model_unc}

Single-factor perturbation ($\pm$10pp): top agent invariant ($U_{\text{model}}{=}0$); mid-ranked swap (Table~\ref{tab:uncertainty}). Dirichlet sampling (1,000 draws from $\mathrm{Dir}(3.5, 2.5, 2.0, 2.0)$): median $\tau{=}0.80$, 90\% CI $[0.63, 0.92]$. \textbf{Concentration sensitivity:} at $k{=}2$ (near-uniform), $\tau{=}0.52$; $k{=}5$, $\tau{=}0.68$; $k{=}10$, $\tau{=}0.80$; $k{=}20$, $\tau{=}0.91$; $k{=}50$, $\tau{=}0.97$. Rankings tolerate substantial weight variation but degrade under near-uniform priors. Bootstrap CIs (1,000 resamples): no top-20 agent shifts by $>{\pm}0.018$.

\section{Validation}
\label{sec:validation}

\textbf{Factor independence ($n{=}50$).} $B$--$A$: $\rho{=}0.05$; $B$--$S$: $\rho{=}0.27$; $A$--$E$: $\rho{=}0.61$ (demand vs.\ supply). Four factors capture largely complementary signal.

\textbf{Predictive validity ($n{=}35$).} B+S sub-composite (no GitHub signals) predicts GitHub stars ($\rho_s{=}0.52$, $p{<}0.01$), SO questions ($\rho_s{=}0.49$, $p{<}0.01$). Leave-one-out: $[0.48, 0.56]$ across all 50 drops.

\textbf{Ablation ($n{=}11$, exploratory).} Full composite vs.\ SWE-bench: $\rho{=}0.03$ (closed-source agents with zero adoption). Power at $n{=}11$: only 0.38; the $n{=}35$ test achieves $>$0.90.

\begin{figure}[!htbp]
\centering
\includegraphics[width=\columnwidth]{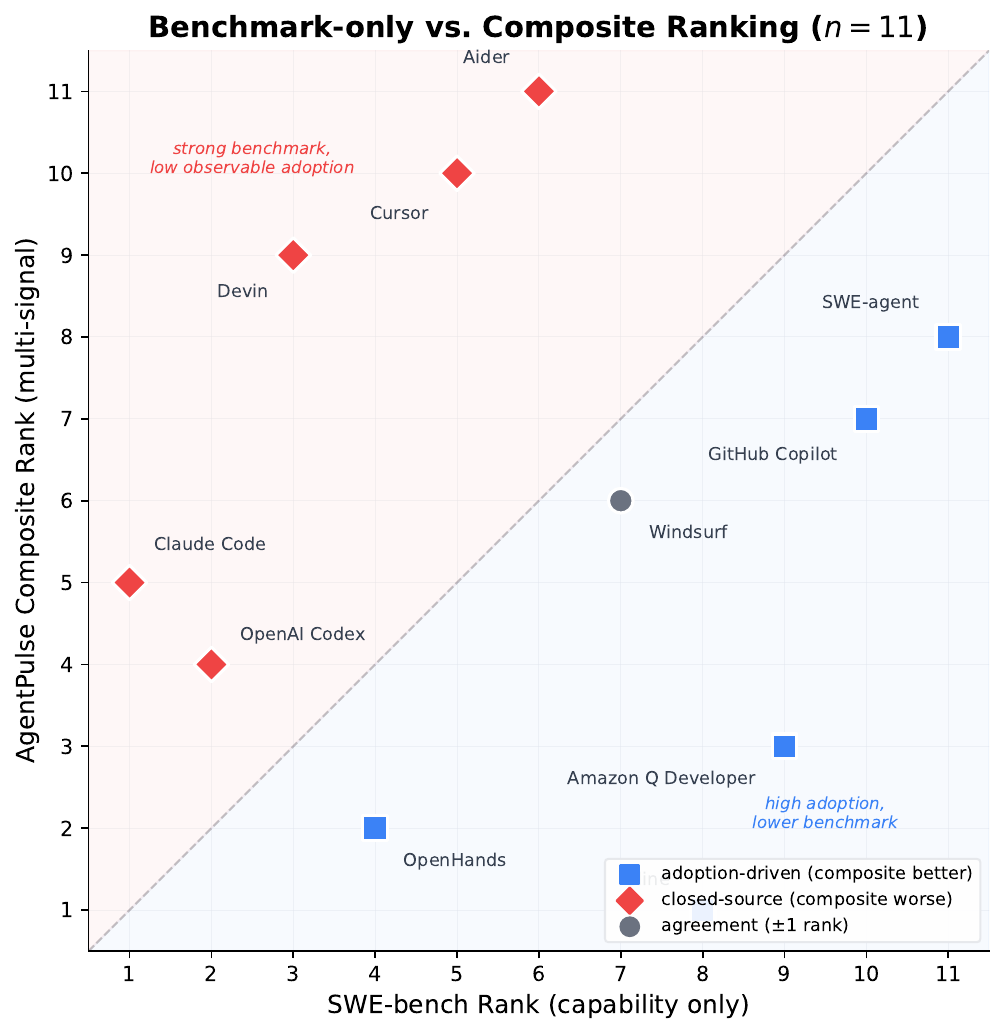}
\caption{Benchmark-only vs.\ composite ranking ($n{=}11$). Blue: adoption-driven; red: closed-source penalty.}
\label{fig:divergence}
\end{figure}

\textbf{Case study.} Cline (0.585) vs.\ OpenHands (0.530): conformal CIs overlap, score-difference interval contains zero, abstention triggered. Claude Code vs.\ Copilot: CIs separated, pair confidently ranked. A newly released agent ($\sigma_{\text{cross}}{=}0.11$) triggered ACI widening; its 72h forecast correctly predicted hype decay.


\section{Discussion}
\label{sec:discussion}

\textbf{What's new about conformal here.} The contribution is not split conformal itself but its application to a domain with four specific challenges: (a) structured exchangeability violations (agent releases) addressed via ACI; (b) conditional coverage failure for volatile agents, addressed via Mondrian conformal with $\sigma_{\text{cross}}$ stratification (lifting volatile coverage from 64.6\% to 80.4\%); (c) compositional structure (pipelines) addressed via validated propagation bounds; (d) leaderboard-scale multiple testing addressed via FDR-controlled abstention. The bootstrap comparison further shows conformal outperforms the field's standard UQ approach under temporal non-stationarity.

\textbf{Future directions.} (1)~Anytime-valid monitoring via e-processes~\citep{ramdas2023gametheoretic}. (2)~Finer-grained Mondrian strata (e.g., by category $\times$ volatility). (3)~Bayesian model averaging over weight priors. (4)~Tighter compositional bounds via empirical pipeline data. (5)~Evaluation uncertainty may proxy for behavioral uncertainty, though we leave empirical validation to future work.

\textbf{Limitations.} (1) ACI requires bounded shift; unbounded regime changes may violate convergence. (2) Compositional bounds are anti-conservative under negative inter-stage correlation (Figure~\ref{fig:compositional}). (3) FDR control assumes independence or positive regression dependence of test statistics across pairs. (4) $n{=}11$ ablation is underpowered. (5) English-only NLP. (6) Closed-source agents lack adoption signals.

\section{Conclusion}

AgentPulse provides distribution-free uncertainty for continuous agent evaluation. Conformal intervals achieve calibration error below 0.02 and outperform bootstrap CIs under non-stationarity. ACI adapts to release-driven distribution shifts. Mondrian conformal lifts volatile-agent coverage from 64.6\% to 80.4\%. Compositional bounds, validated by simulation, propagate uncertainty through pipelines. FDR-corrected abstention controls leaderboard-scale false-ranking rate. Code and data are released under CC BY 4.0.

\bibliography{agentpulse_refs}
\bibliographystyle{icml2026}

\end{document}